\documentclass[10pt,twocolumn,letterpaper]{article}
\usepackage{iccv}
\usepackage{times}
\usepackage{epsfig}
\usepackage{graphicx}
\usepackage{amsmath}
\usepackage{amssymb}
\usepackage{float} 
\usepackage[numbers]{natbib}
\usepackage{fancyhdr}

\usepackage[pagebackref=true,breaklinks=true,letterpaper=true,colorlinks,bookmarks=false]{hyperref}

\iccvfinalcopy 

\def\iccvPaperID{11} 

\ificcvfinal\fi
\usepackage{xcolor}

\fancypagestyle{plain}{%
	\fancyhf{}%
	\fancyfoot[L]{\footnotesize \center \emph{Preprint accepted for publication in the Proceedings of the 2nd International Workshop on Computer Vision for Physiological Measurement as part of ICCV. Seoul, South Korea. 2019. \\ -\thepage-}}%
}

\begin{document}
\pagestyle{fancy}

\title{Multimodal Deep Models for Predicting Affective Responses Evoked by Movies}

\author{Ha Thi Phuong Thao \hspace*{1cm} Dorien Herremans \hspace*{1cm} Gemma Roig\\
Singapore University of Technology and Design\\
8 Somapah Rd, Singapore 487372\\
{\tt\small thiphuongthao\_ha@mymail.sutd.edu.sg, dorien\_herremans@sutd.edu.sg,  gemmar@mit.edu}
}

\maketitle
\cfoot{\footnotesize \center \emph{Preprint accepted for publication in the Proceedings of the 2nd International Workshop on Computer Vision for Physiological Measurement as part of ICCV. Seoul, South Korea. 2019. \\ -\thepage-}}   

\begin{abstract}
   The goal of this study is to develop and analyze multimodal models for predicting experienced affective responses of viewers watching movie clips. We develop hybrid multimodal prediction models based on both the video and audio of the clips.  For the video content, we hypothesize that both image content and motion are crucial features for evoked emotion prediction. To capture such information, we extract features from  RGB frames and optical flow using pre-trained neural networks. For the audio model, we compute an enhanced set of low-level descriptors including intensity, loudness, cepstrum, linear predictor coefficients, pitch and voice quality. Both visual and audio features are then concatenated to create audio-visual features, which are used to predict the evoked emotion. To classify the movie clips into the corresponding affective response categories, we propose two approaches based on deep neural network models. The first one is based on fully connected layers without memory on the time component, the second incorporates the sequential dependency with a long short-term memory recurrent neural network (LSTM). We perform a thorough analysis of the importance of each feature set. Our experiments reveal that in our set-up, predicting emotions at each time step independently gives slightly better accuracy performance than with the LSTM. Interestingly, we also observe that the optical flow is more informative than the RGB in videos, and overall, models using audio features are more accurate than those based on video features when making the final prediction of evoked emotions.
\end{abstract}

\section{Introduction} \label{sec:intro}
Human emotional experiences can be evoked by watching audio-visual content, such as movies.  In psychology, evoked emotions have been extensively studied~\cite{zentner2008emotions, koelsch2010towards, baumgartner2006emotion, gabrielsson2001emotion}. Being able to automatically predict which emotions multimedia content might evoke has a wide range of applications. For instance, it can be a tool for multimedia producers in advertisement or film industry.  Yet, in computer science, most of the current and previous research focuses on emotion recognition of people in videos, and they are based on facial expressions and audio signals~\cite{kahou2016emonets, hu2017learning, ebrahimi2015recurrent}. Predicting the viewers' emotion  evoked by videos and multimedia content has received little attention so far.  

For measuring  affective responses, some researchers have proposed to use emotional categories and suggested that the number of distinct emotional categories may vary from two to twenty seven~\cite{cowen2017self, picard2000affective}. Although the precise categories can be different in studies, the two most common ones are arousal and valence, which were originally proposed in ~\cite{russell1980circumplex}, and have been used in most predictive models,  \eg~\cite{samara2016feature, zlatintsi2017cognimuse,herremans2017imma, lang1995emotion}. Arousal ranges from calm to exciting, while valence represents how positive or negative the emotion is~\cite{picard2000affective}. \citet{russell1980circumplex} also proposed another factor, namely dominance, which refers to the sense of ``control'' or ``attention'' ~\cite{picard2000affective}. Dominance, however, has been known to introduce complexity in the annotation process and is difficult to computationally predict~\cite{zlatintsi2017cognimuse}, hence, it is often omitted. 
\begin{figure}[t!]
\begin{center}
\includegraphics[width=1.0 \linewidth]{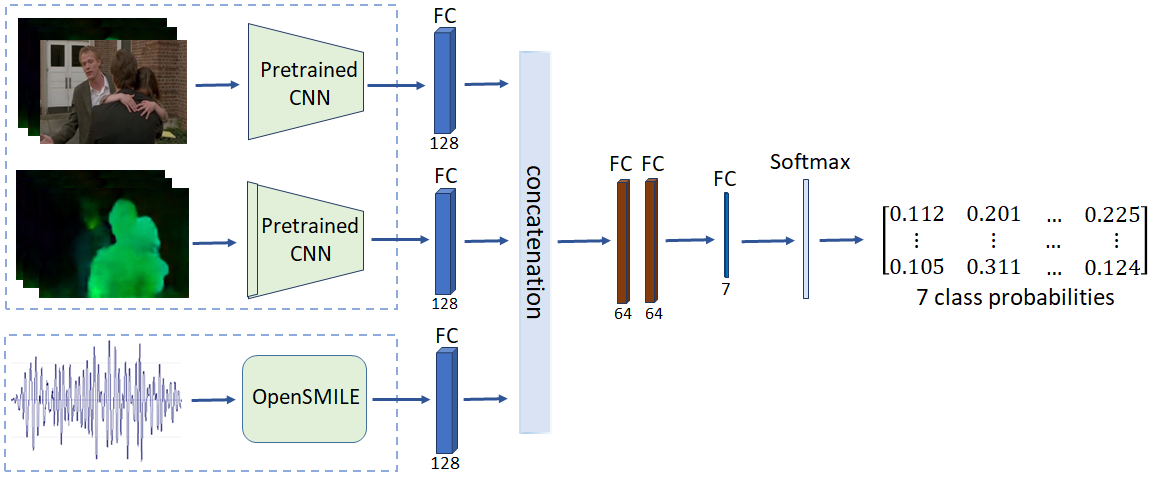}  
\end{center}
   \caption{Our proposed multimodal architecture for emotion classification using fully connected layers.}
\label{fig:RGB_OF_Audio_2FC_upgraded}
\end{figure}

In this work, we propose a model for predicting evoked emotion from videos. We use the two-dimensional model of affective content, in which arousal and valence are predicted separately. For doing so, we  leverage  the power of deep convolutional neural networks (CNNs), a type of network that has led to considerable advances in image classification~\cite{krizhevsky2012imagenet, simonyan2014very, he2016deep, szegedy2015going} and action recognition~\cite{ji20123d, simonyan2014two, carreira2017quo}. We apply deep CNNs for extracting image and motion features from static RGB frames and optical flow fields respectively. This approach is similar to the two-stream ConvNet architecture in \cite{simonyan2014two}, in which spatial and temporal networks are integrated for action recognition. The spatial network is used to capture features from scenes and objects in videos, while the temporal network carries information about the motion of camera and objects across frames. The audio features are computed using OpenSMILE~\cite{EybenOpenSMILE}, and include intensity, loudness, cepstrum, linear predictor coefficients, pitch, voice quality, among others.

We explore two models based on deep neural networks that ingest the extracted features. The first one consists of fully connected layers without memory on the time component, as shown in Figure~\ref{fig:RGB_OF_Audio_2FC_upgraded}. The second model that we explore uses long-short term memory (LSTM) structures, which are recurrent neural networks with memory that can learn temporal dependencies between observations. LSTMs have been successfully applied in sequence prediction problems including emotion recognition of subjects in a video~\cite{fan2016video,  pini2017modeling}.

We perform experiments on the extended COGNIMUSE dataset~\cite{malandrakis2011supervised}, which consists of 12 movie clips: seven half-hour continuous movie clips from the original COGNIMUSE dataset~\cite{evangelopoulos2013multimodal} and five extra half-hour Hollywood movie clips. Valence and arousal are annotated in the range of~$\left[-1, 1\right]$ by several subjects. We perform a throughout analysis of the importance of the video and audio feature set. Our results suggest that audio contains most of the evoked emotion content, and for videos, motion is more important than the RGB frames. We also observe that our model with fully connected layers outperforms the model with an LSTM structure, as well as a previous approach introduced with the extended COGNIMUSE dataset~\cite{malandrakis2011supervised}. 
%

\section{Related work} 
\label{sec:Related_work}
In computer vision, emotion prediction from video has been mostly studied from the perspective of predicting facial expressions of humans in the videos, \cf~\cite{Kanade:2005:FEA:2101315.2101317,Cohn_automatedface}. Yet, predicting evoked emotion has received surprisingly little attention so far.

In most current research in video-based emotion recognition~\cite{levi2015emotion, fan2016video, kaya2017video, zheng2018multimodal, kahou2016emonets}, multimodal approaches have been applied to integrate information from different modalities. The recent breakthrough of deep convolutional neural networks (CNNs) for object recognition~\cite{krizhevsky2012imagenet, simonyan2014very, he2016deep} has been adapted to the problem of emotion recognition in videos, in which deep CNNs such as VGG~\cite{simonyan2014very} and Inception-ResNet v1~\cite{szegedy2017inception} are used to extract facial expression features from RGB frames ~\cite{fan2016video, zheng2018multimodal, pini2017modeling}. In addition to features extracted from still frames, motion plays an important role in emotion recognition. Actions represented by facial muscles or body actions in videos may be estimated using optical flow~\cite{mase1991recognition, simonyan2014two}. \citet{mase1991recognition} recognizes four basic facial expressions including surprise, happiness, disgust and anger based on optical flows. The appearance and action information can be handled separately in two streams using CNNs~\cite{simonyan2014two, wang2015towards} or extracted simultaneously using deep three-dimensional convolutional networks~\cite{fan2016video, pini2017modeling}.   

Many studies have shown that there is a powerful connection between sound and emotion~\cite{Meyer56, panksepp2002emotional, zentner2008emotions, doughty2016practices, herremans2016}, therefore, it is necessary to add the audio modality into emotion recognition models. Audio features can be extracted using toolkits such as OpenSMILE~\cite{EybenOpenSMILE}, YAAFE~\cite{mathieu2010yaafe} or deep neural networks such as SoundNet~\cite{aytar2016soundnet}, AlexNet, Inception, ResNet architectures~\cite{hershey2017cnn}. In this study, we opted for the OpenSMILE toolkit as it has shown its efficiency in emotion recognition models from voice.   

Another important aspect for predicting evoked emotion from movies might be the temporal component. To deal with sequences, recurrent neural networks have been successfully used in many applications. However, they are limited by their long-term dependency, where the state information is integrated over time resulting in gradient exploding/vanishing when training them to learn long-term dynamics. In order to overcome this limitation, the LSTM cell was first introduced by~\citet{hochreiter1997long} and later simplified in~\cite{graves2013generating, zaremba2014learning}. LSTMs have a good performance in a wide range of sequence processing research and have also been widely used in video-based emotion recognition~\cite{fan2016video, pini2017modeling, zheng2018multimodal}. 

Most previous work focuses on predicting the emotion of humans in a video instead of evoked/experienced emotion from viewers of videos. This is in part due to the lack of available labeled datasets. Recently, to fill this gap, \citet{zlatintsi2017cognimuse} introduced the COGNIMUSE dataset, which is a multimodal video dataset including seven half-hour Hollywood movie clips.  \citet{malandrakis2011supervised} use the extended version of this dataset, which includes 12 movies clips and classify emotion in terms of seven valence and arousal categories at the frame level using independent hidden Markov models (HMMs). A wide range of visual and audio features are extracted, but finally only a small feature set including mel-frequency cepstral coefficients (MFCCs), their derivatives, maximum color value and maximum color intensity is selected. A follow up approach including a mixture of expert models to select the audio and video features dynamically is introduced by~\citet{goyal2016multimodal}, and~\citet{sivaprasad2018multimodal}, who improve their results by using the LSTM structure, yet they predict valence and arousal only every 5 seconds instead of at every frame. In~\cite{malandrakis2011supervised,zlatintsi2017cognimuse}, instead of predicting continuous values of affective content, they discretize those values in equally spaced ranges, and assign a label to each of them. They show that predicting labels and then interpolate them to continuous values gives better results than regressing the emotion values. 
 
In this work, we use the extended COGNIMUSE dataset to learn multimodal models for predicting evoked/experienced emotion, in terms of valence and arousal, from videos. Interestingly, this dataset also contains video frames in which people do not appear, hence previous approaches based on facial expression recognition cannot be applied. In the following, we elaborate on our approach and describe our hybrid multimodal model, which is based on deep neural networks. In the result section, we provide an analysis of the features and components that contribute most to the prediction accuracy. 
%

\section{Approach}
\label{sec:Approach}
\begin{figure}
\begin{center}
\includegraphics[width=1.0 \linewidth]{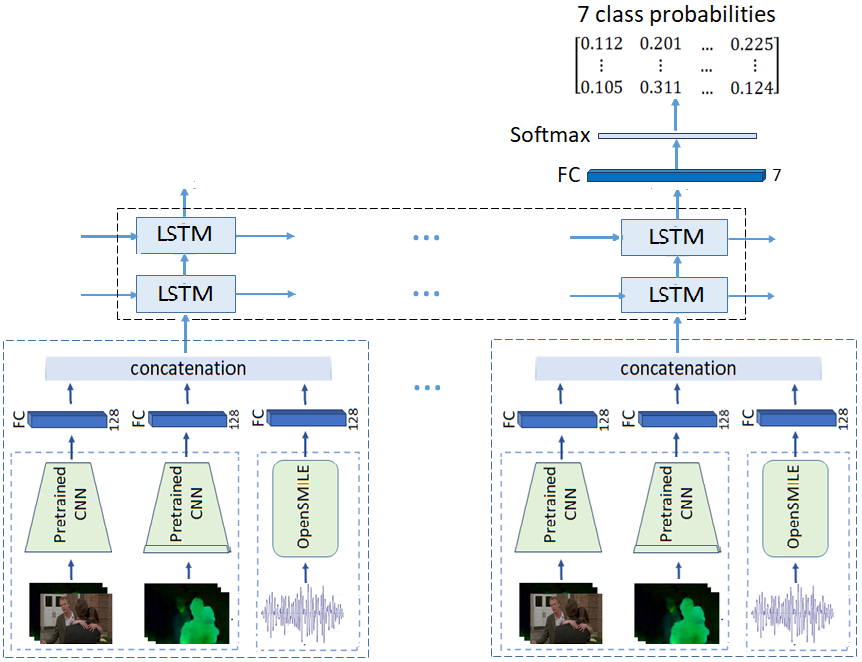}
\end{center}
   \caption{A multimodal architecture for emotion classification using an LSTM structure.}
\label{fig:RGB_OF_Audio_LSTM}
\end{figure}

We propose a multimodal approach that uses both video and audio features for emotion prediction. For the former, we use pre-trained CNNs to extract image features from static RGB frames and motion features from optical flow fields. The latter is based on features from the OpenSMILE toolkit~\cite{EybenOpenSMILE}. Each of these feature sets are passed through fully connected layers for dimensionality reduction and representation adaptation to emotion prediction, before being concatenated to create audio-visual features. The weights of these fully connected layers are learned jointly with our proposed network architecture during training. We explore two network architectures. The first model uses fully connected layers without temporal memory component (Figure~\ref{fig:RGB_OF_Audio_2FC_upgraded}), while the second one is based on LSTMs to take the sequential dependency of emotion into account (Figure~\ref{fig:RGB_OF_Audio_LSTM}). Both of these two approaches are followed by a fully connected layer and a softmax layer to classify arousal and valence separately, as it has been shown that those are orthogonal emotion characteristics~\cite{russell1980circumplex}, and their accuracy prediction suffers when being estimated jointly. Details on each of the components of our proposed models are further discussed below. In the experimental section, we report the results of an analysis of the importance of each feature for evoked emotion prediction in videos. The code to reproduce our results and with the models' implementation is available at: \url{https://github.com/ivyha010/emotionprediction}.

\subsection{Visual feature extraction} 
\label{subsec:Video}
We extract spatial features from the static RGB frames and motion cues from the optical flow of consecutive frames, similarly to the two-stream ConvNet approach in \cite{simonyan2014two}. The spatial component provides information about objects and scenes in single still RGB frames. The temporal component contains information about motion. Each of these components are extracted using pre-trained CNN. 
This approach has similarities to the two-stream ConvNets used for action recognition, as those also use both optical flow and still RGB frames as input~\cite{simonyan2014two}. Yet, in~\cite{simonyan2014two}, each of the CNN streams is relatively shallow in comparison with  CNNs trained on ImageNet~\cite{he2016deep,simonyan2014very,krizhevsky2012imagenet}. It is advantageous for us to use a pre-trained CNN as our dataset is relatively small.
\paragraph{Semantic content from still frames}
For the spatial component, image features from still RGB frames are extracted using a CNN with pre-trained weights on the ImageNet dataset for object classification task, namely ResNet-50~\cite{he2016deep}.  We extract the representation from the second-to-last layer, \ie after forward passing the image to all the layers except for the last fully-connected classification layer.
\paragraph{Optical Flow}
In our framework, we estimate a dense optical flow using PWC-Net~\cite{sun2018pwc} pre-trained on MPI Sintel final pass dataset~\cite{butler2012naturalistic}. It is a CNN model designed according to three principles: pyramid processing, warping, and the application of a cost volume. It computes the optical flow fields between pairs of successive frames. 
We use PWC-Net since it has a smaller size than FlowNet2.0~\cite{ilg2017flownet}, which makes it easier to train, however, it outperforms many other dense optical flow methods such as SPyNet~\cite{ranjan2017optical}, DC Flow~\cite{xu2017accurate} and flow fields ~\cite{bailer2015flow} on MPI Sintel final pass.

The estimated optical flow fields are transformed into integers in [0, 255] as in~\cite{wang2015towards} to store them in two channels of JPEG images, values in the third channel are set to 255. 
We use a stack of 10 sequential optical flows, as it carries more motion information than the optical flow between two consecutive frames, as suggested in~\cite{wang2015towards}. The stack serves as the input to the ResNet-101 model that has been pre-trained on the ImageNet classification task, except for the first convolutional layer and the last classification layer, which had been fine-tuned to be able to ingest 10 stacks of sequential optical flows to predict action recognition on UCF-101~\cite{wang2015towards}\footnote{Pre-trained model available at: https://github.com/jeffreyhuang1/two-stream-action-recognition}.  We remove the last fully connected classification layer in the ResNet-101 model and freeze the rest. We thus extract a 2,048-dimensional feature vector from every stack of 10 optical flows.   

\subsection{Audio feature extraction} \label{subsec:Audio}
The audio present in movie clips typically consists of a combination of speech, music, and  sound effects meant to engage the audience in the stories that filmmakers want to deliver. In our proposed system, audio features are extracted using the OpenSMILE toolkit with a frame size of 400ms with a hop size of 40ms. The frame size corresponds to a time period of a stack of 10 optical flows as shown in Figure \ref{fig:Frame_OF_Audio_block}. We use a configuration file named ``{\em emobase2010}'', which is based on INTERSPEECH 2010 paralinguistics challenge~\cite{schuller2010interspeech} to extract 1,582 features including low-level descriptors (\eg pitch, loudness, jitter, MFCCs, mel filter-bank, line spectral pairs) with their delta coefficients, functionals, the number of pitch onsets, and the duration in seconds~\cite{eyben2016open}. This set of audio features is relatively large in comparison to those created by other OpenSMILE configuration files.  

\subsection{Fusion of extracted features}
\begin{figure}
\begin{center}
\includegraphics[width=1.0 \linewidth]{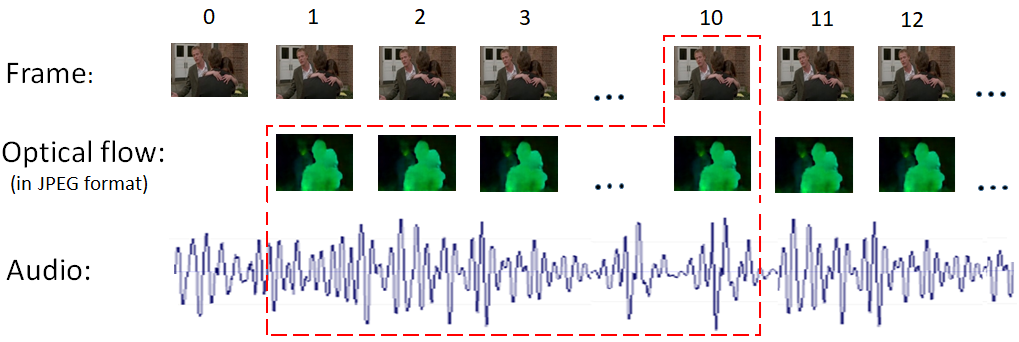}
\end{center}
   \caption{The three types of input used: image, optical flow stack, and audio segment.}
\label{fig:Frame_OF_Audio_block}
\end{figure}

We analyze the effect of multimodal inputs including image features, motion features and audio features on predicting evoked emotion that viewers actually encounter when watching movie clips. Image features are extracted from single RGB frames while motion features and audio features come from 10-optical flow stacks and 400ms-audio segments respectively, as shown in Figure~\ref{fig:Frame_OF_Audio_block}. 
Each extracted feature is normalized using min-max normalization following this formula:  $\text{V}_{i}^{norm} = \frac{\text{V}_{i}-\min \left(\textbf{V}\right) }{\max \left(\textbf{V}\right)-\min \left(\textbf{V}\right)}$, in which  $\text{V}_{i}$ is the $i$-th data point in vector $\textbf{V}$ that contains the same feature element for all data points. 

In order to reduce the dimension of the extracted feature vectors, we pass the extracted features of each modality to a fully connected layer of 128 units as showed in Figures \ref{fig:RGB_OF_Audio_2FC_upgraded} and~\ref{fig:RGB_OF_Audio_LSTM}. The weights of this layer are learned during training and optimized for predicting emotion. Then, the output of these fully connected layers are concatenated before being fed into another two fully connected layers as described in Figure \ref{fig:RGB_OF_Audio_2FC_upgraded} or the LSTM structure in Figure~\ref{fig:RGB_OF_Audio_LSTM}.   
%
\subsection{Models for emotion recognition} \label{subsec:Models}
We propose two variants of the model for emotion classification. The first one includes only fully connected layers without memory on the time component, while the second one takes the sequential dependency of emotion responses into account by using a LSTM structure. Each of these models are created for arousal and valence separately. Both of these approaches are followed by a fully connected layer and a softmax layer to classify arousal and valence. Since valence and arousal are real values in the range $[-1,1]$, we convert the prediction problem into classification problem by quantizing the real values into 7 bins. In this way, we  are be able to use the cross-entropy loss, which gives better results in practice than optimizing the mean squared error loss. The same binning has been performed in~\cite{malandrakis2011supervised}, which allows us to benchmark our results. %
\paragraph{Model with no sequential memory}
The audio-visual features are fed into two fully connected layers after the fusion of extracted features. We use 64 units per layer, as described in Figure~\ref{fig:RGB_OF_Audio_2FC_upgraded}. The outputs of the two fully connected layers are then passed to a smaller fully connected layer consisting of 7 units followed by a softmax layer that provides the final probability output for each of the seven binned emotion responses.
%
%
\paragraph{Model with sequential memory}
 We implemented an LSTM in order to incorporate the time dependencies when predicting the affective response of viewers watching movies. The basic architecture of an LSTM cell includes a cell $c_{t}$, remembering values over time and three gates: input gate $i_{t}$, forget gate $f_{t}$, and output gate $o_{t}$. The LSTM cell can be described using the following equations~\cite{donahue2015long}: 
\begin{align}
f_{t} &= \textit{sigmoid} \left( W_{xf} x_{t} + W_{hf} h_{t-1} + b_{f} \right) \label{eq:LSTM-cell-1} \\
i_{t} &= \textit{sigmoid} \left( W_{xi} x_{t} + W_{hi} h_{t-1} + b_{i} \right) \label{eq:LSTM-cell-2} \\
g_{t} &= \textit{tanh} \left( W_{xc} x_{t} + W_{hc} h_{t-1} + b_{c} \right) \label{eq:LSTM-cell-3} \\
c_{t} &= f_{t} \odot c_{t-1} + i_{t} \odot g_{t} \label{eq:LSTM-cell-4} \\
o_{t} &= \textit{sigmoid} \left( W_{xo} x_{t} + W_{ho} h_{t-1} + b_{o} \right) \label{eq:LSTM-cell-5} \\
h_{t} &= o_{t} \odot \textit{tanh} \left( c_{t} \right) \label{eq:LSTM-cell-6}
\end{align}
in which $x_{t}$ is the input ($t = 1, \dots, T$),  $T$ is the input sequence length, $h_{t} \in \mathbb{R}^{N}$ is the hidden state with $N$ being the number of hidden units; $ W_{xf}$, $ W_{hf}$, $W_{xi}$, $W_{hi}$, $W_{xc}$, $W_{hc}$, $W_{xo}$ and $W_{ho}$ are matrices of weights; $b_{f}$, $b_{i}$, $b_{c}$ and $b_{o}$ are biases; $\textit{sigmoid}$ is the sigmoid function $sigmoid (x) = \frac{1}{1+e^{-x}}$; $\odot$ is element-wise product.
The forget gate is the first and most important gate, which resets the LSTM cell state using a sigmoid function (Equation~\ref{eq:LSTM-cell-1}). The input gate decides which values will be updated using a sigmoid function  (Equation~\ref{eq:LSTM-cell-2}), and a $\tanh$ function is used to create a vector $g_{t}$ of new updated values (Equation~\ref{eq:LSTM-cell-3}). The cell state is computed from the forget gate, the previous cell state, input gate and the vector of new updated values (Equation~\ref{eq:LSTM-cell-4}). At the output gate, a sigmoid function is used to decide which part of the cell state is going to be the final output (Equation~\ref{eq:LSTM-cell-5}). The cell state is put through a $\tanh$ function to convert the values into the range $\left[ -1, 1 \right]$ and multiplied by the output (Equation~\ref{eq:LSTM-cell-6}). 

In our network, we use a two-layer LSTM structure, each with a hidden size of 64 units. The LSTM model works on overlapping input sequences, which are sequences of audio-visual feature vectors, and provides only one output for each sequence of inputs. We use a sequence length of $5$ time steps, which is equivalent to 2 seconds. The 64-dimensional output of the last time step of the LSTM model is passed through a fully-connected layer of seven units followed by a softmax layer as shown in Figure $\ref{fig:RGB_OF_Audio_LSTM}$. 

%
\section{Experimental Set-up} \label{sec:Experimental_Setup}
We detail the dataset and the implementation of our experimental set-up in what follows. 
\paragraph{Dataset} We report results on the extended COGNIMUSE dataset~\cite{malandrakis2011supervised}, which consists of seven half-hour continuous movie clips from the COGNIMUSE dataset~\cite{evangelopoulos2013multimodal} and five half-hour additional Hollywood movie clips. This dataset includes annotation for sensory and semantic saliency, events, cross-media semantics and emotion, however, in this study we focus on the emotion annotation. Emotion is represented in continuous arousal and valence values in the range $\left[ -1, 1 \right]$. There are three types of annotated emotions: intended, expected and experienced emotion~\cite{zlatintsi2017cognimuse}, however, expected emotion is computed
from the experienced emotion annotations, therefore, only intended and experienced emotions are rated in this dataset. We focus mainly on experienced emotion, which is equivalent to the evoked emotion, and described in terms of valence and arousal values computed as the average of twelve annotations. To be able to compare to previous work, we also report results on intended emotions, which represent the intention of film makers, and are also annotated in terms of valence and arousal values, computed as the average of three annotations done by the same expert at frame level. In both cases, the  emotion values (valence and arousal), which range between $-1$ and $1$, are quantized into seven bins as suggested in~\cite{malandrakis2011supervised}. This enables us to tackle the problem as a labeling task, which  results in  better results.
\paragraph{Data pre-processing} The movie clips all have a frame rate of 25 frames per second, but vary in frame resolution. Seven movies in the dataset have a height under $214$ pixels, therefore, we resize their raw RGB video frames to meet the  size requirement of $224$ for each dimension of images to be fed into the ResNet-50 pre-trained on ImageNet. For movies with larger frame size, we take a random crop of a $224 \times 224$ region. In all cases, we scale the RGB channels by substracting the mean and dividing by the standard deviation of the RGB frames from the ImageNet dataset. For the optical flow network, we keep the original size of the RGB frames as the input, and rescale the optical flow outputs to match the size of $224 \times 224$ required by the ResNet-101.
\paragraph{Evaluation metrics}
The proposed models are evaluated based on leave-one-out cross-validation, in which the accuracy and accuracy$\pm 1$ (i.e, predictions of the class label adjoined to the real class is also considered as correct) are used for emotion classfication. We refer to this evaluation as the ''discrete case". We also compute the mean absolute error (MAE), mean squared error (MSE) and Pearson correlation coefficient with respect to the ground truth by converting the discrete predicted outputs of valence and arousal to continuous values (i.e., ''the continuous case" in the tables below). This is done by following Malandrakis' approach~\cite{malandrakis2011supervised}, in which a low pass filter is applied on classification outputs to eliminate the noise before using the Savitzky-Golay filter~\cite{savitzky1964smoothing} and rescaling into the range $\left[-1,1 \right]$. 
\paragraph{Implementation details}
The models that classify valence and arousal separately into seven classes are trained using stochastic gradient descent (SGD) with a learning rate of $0.005$, a weight decay of $0.005$, and the softmax function with a temperature of $T = 2$. We train the models for $200$ epochs, each with a batch size of $128$ and early stopping with a patience of 25 epochs. For the LSTM, we set a fixed sequence length equal to 5. All the models are implemented in Python 3.6 and the experiments were run on a NVIDIA GTX 1070. 
%
%
%
\begin{table*}[t!]
\begin{center}
\begin{tabular}{l|c|c|c|c|c|}
\cline{2-6}
                       & \multicolumn{2}{c|}{Discrete case} & \multicolumn{3}{c|}{Continuous case} \\ \cline{2-6} 
                       & Accuracy (\%) & Accuracy $\pm$ 1 (\%) & MAE & MSE &  Correlation \\ \hline
\multicolumn{1}{|l|}{\textbf{Model with FC layers}} & & & & & \\ 
\multicolumn{1}{|l|}{RGB frame} & 49.04 & 92.84 & 0.17 & 0.05 &  0.31\\ 
\multicolumn{1}{|l|}{Optical Flow (OF)} & 51.08 & 93.90 & 0.18 & 0.05 & 0.34 \\ 
\multicolumn{1}{|l|}{Audio} & 51.10 & 95.67 & 0.15 & 0.04 & 0.44 \\ 
\multicolumn{1}{|l|}{RGB frame + OF + Audio} & \textbf{53.32} & \textbf{94.75} & \textbf{0.15} & \textbf{0.04} & \textbf{0.46} \\ \hline
\multicolumn{1}{|l|}{\textbf{Model with LSTM}} & & &  & &  \\ 
\multicolumn{1}{|l|}{RGB frame + OF + Audio} &  48.64 & 95.28 & 0.37 & 0.17 & 0.43   \\ \hline
\end{tabular}
\end{center}
\caption{Leave-one-out cross-validation results for the arousal dimension in experienced emotion using different input modalities.}
\label{tab:Arousal_results_Experienced}
\end{table*}
%
%
%
\begin{table*}[t!]
\begin{center}
\begin{tabular}{l|c|c|c|c|c|}
\cline{2-6}
                       & \multicolumn{2}{c|}{Discrete case} & \multicolumn{3}{c|}{Continuous case} \\ \cline{2-6} 
                       & Accuracy (\%) & Accuracy $\pm$ 1 (\%) & MAE & MSE &  Correlation \\ \hline
\multicolumn{1}{|l|}{\textbf{Model with FC layers}} & & & & & \\ 
\multicolumn{1}{|l|}{RGB frame} & 38.60 & 90.24 &  0.20 & 0.06 & 0.05 \\ 
\multicolumn{1}{|l|}{Optical Flow (OF)} & 42.35 & 90.12 & 0.19 & 0.06 & 0.15  \\ 
\multicolumn{1}{|l|}{Audio} & 42.53 & 89.01 & 0.19 & 0.06 & 0.15 \\ 
\multicolumn{1}{|l|}{RGB frame + OF + Audio} & \textbf{43.10} & \textbf{90.51} & \textbf{0.19} & \textbf{ 0.06} & \textbf{0.18} \\ \hline
\multicolumn{1}{|l|}{\textbf{Model with LSTM}} & & &  & &  \\ 
\multicolumn{1}{|l|}{RGB frame + OF + Audio} & 37.20 & 89.22 & 0.22 & 0.07 & 0.05  \\ \hline
\end{tabular}
\end{center}
\caption{Leave-one-out cross-validation for the valence dimension in experienced emotion using different input modalities.}
\label{tab:Valence_results_Experienced}
\end{table*}
%
%
\begin{table*}[t!]
\begin{center}
\begin{tabular}{l|c|c|c|c|c|}
\cline{2-6}
                       & \multicolumn{2}{c|}{Discrete case} & \multicolumn{3}{c|}{Continuous case} \\ \cline{2-6} 
                       & Accuracy (\%) & Accuracy $\pm$ 1 (\%) & MAE & MSE &  Correlation \\ \hline
\multicolumn{1}{|l|}{\textbf{Model with FC layers}} & & & & & \\ 
\multicolumn{1}{|l|}{RGB frame} & 27.63 & 64.89 & 0.34& 0.20 & 0.42\\ 
\multicolumn{1}{|l|}{Optical Flow (OF)} & 28.39 & 66.89 & 0.35 & 0.21 & 0.46 \\ 
\multicolumn{1}{|l|}{RGB frame + OF} & 28.98 & 66.43 & 0.35 & 0.21 & 0.40 \\ 
\multicolumn{1}{|l|}{Audio} & 30.81 & 72.90 & 0.28 & 0.13 & 0.59 \\ 
\multicolumn{1}{|l|}{RGB frame + OF + Audio} & \textbf{31.20} & \textbf{72.94} & \textbf{0.27} & \textbf{0.13} & \textbf{0.62} \\ \hline
\multicolumn{1}{|l|}{\textbf{Model with LSTM}} & & &  & &  \\ 
\multicolumn{1}{|l|}{RGB frame + OF + Audio} & 30.80 & 71.69 & 0.41 & 0.22 & 0.58 \\ \hline
\multicolumn{1}{|l|}{\citet{malandrakis2011supervised}} & 24.00 & 57.00 & 0.32 &0.17 & 0.54 \\ \hline
\end{tabular}
\end{center}
\caption{Leave-one-out cross-validation for the arousal dimension in intended emotion using different input modalities.}
\label{tab:Arousal_results_Intended}
\end{table*}
%
\begin{table*}[t!]
\begin{center}
\begin{tabular}{l|c|c|c|c|c|}
\cline{2-6}
                       & \multicolumn{2}{c|}{Discrete case} & \multicolumn{3}{c|}{Continuous case} \\ \cline{2-6} 
                       & Accuracy (\%) & Accuracy $\pm$ 1 (\%) & MAE & MSE &  Correlation \\ \hline
\multicolumn{1}{|l|}{\textbf{Model with FC layers}} & & & & & \\ 
\multicolumn{1}{|l|}{RGB frame} & 24.87 & 59.23 & 0.35 & 0.18 & 0.27 \\ 
\multicolumn{1}{|l|}{Optical Flow (OF)} &26.75 &59.36 &0.38 &0.24 & 0.21\\ 
\multicolumn{1}{|l|}{RGB frame + OF} & 24.54 & 56.28 & 0.37 & 0.21 & 0.16 \\ 
\multicolumn{1}{|l|}{Audio} & 29.53 & 65.56 & 0.33 & 0.19 & 0.20\\ 
\multicolumn{1}{|l|}{RGB frame + OF + Audio} & \textbf{30.33} & \textbf{66.95} & \textbf{0.32} & \textbf{0.19} & \textbf{0.25} \\ \hline
\multicolumn{1}{|l|}{\textbf{Model with LSTM}} & & &  & &  \\ 
\multicolumn{1}{|l|}{RGB frame + OF + Audio} & 22.54 & 57.63 & 0.44 & 0.26 & 0.17
\\ \hline
\multicolumn{1}{|l|}{\citet{malandrakis2011supervised}} & 24.00 & 64 & 0.37 &0.24 &0.23  \\ \hline
\end{tabular}
\end{center}
\caption{Leave-one-out cross-validation for the valence dimension in intended emotion using different input modalities.}
\label{tab:Valence_results_Intended}
\end{table*}
\section{Experimental Results} \label{sec:results}
The models are trained and validated on the experienced and intended emotion annotations in the extended COGNIMUSE dataset. The results are summarised in Table~\ref{tab:Arousal_results_Experienced} and Table~\ref{tab:Valence_results_Experienced} for experienced emotion prediction, and in Table \ref{tab:Arousal_results_Intended} and Table \ref{tab:Valence_results_Intended} for intended emotion prediction. 
\paragraph{Analysis of the importance of each  audio-visual feature component} We analyze the effect of the different modalities on classifying emotion of viewers. We use the network with the fully connected layers for this analysis (Figure~\ref{fig:RGB_OF_Audio_2FC_upgraded}). The architecture is kept the same, varying only the input using only one of the following features: RGB frame features, optical flow features, audio, and combinations of those.
%
%
 We observe in Tables~\ref{tab:Arousal_results_Experienced} and~\ref{tab:Valence_results_Experienced} that models based on audio features have a higher classification accuracy than other modalities (image and motion), when predicting experienced emotion.
This may indicate that either the audio features have a larger influence on emotions than visual features, or the audio features used are better suited for emotion prediction than the video features. In fact, the extended Cognimuse dataset includes famous Hollywood movie clips, and hence, speech, sound effects as well as music are used by filmmakers to describe the inner thoughts of characters in movies and deliver some messages to audience. By using a fusion of all feature modalities, we are able to reach the highest performance, slightly improving the results with only audio features, for both arousal and valence classification. 

Similar conclusions can be drawn for predictions of intended emotion (Tables~\ref{tab:Arousal_results_Intended} and~\ref{tab:Valence_results_Intended}). Yet, we observe that for both models, fully connected layers and LSTM structure, the accuracy for predicting experienced emotion is higher than that of intended emotion. However, the Pearson correlation between the predicted and ground-truth experienced valence is lower than in the case of intended emotion. This corresponds with the inter-annotator agreement statistics shown in Table~$11$ in~\cite{zlatintsi2017cognimuse}, which indicate that individual experienced annotations are highly subjective and can vary between annotators, whereas intended annotations are of one subject only.
\begin{figure}[t!]    
\begin{center}
   \includegraphics[scale=0.5]{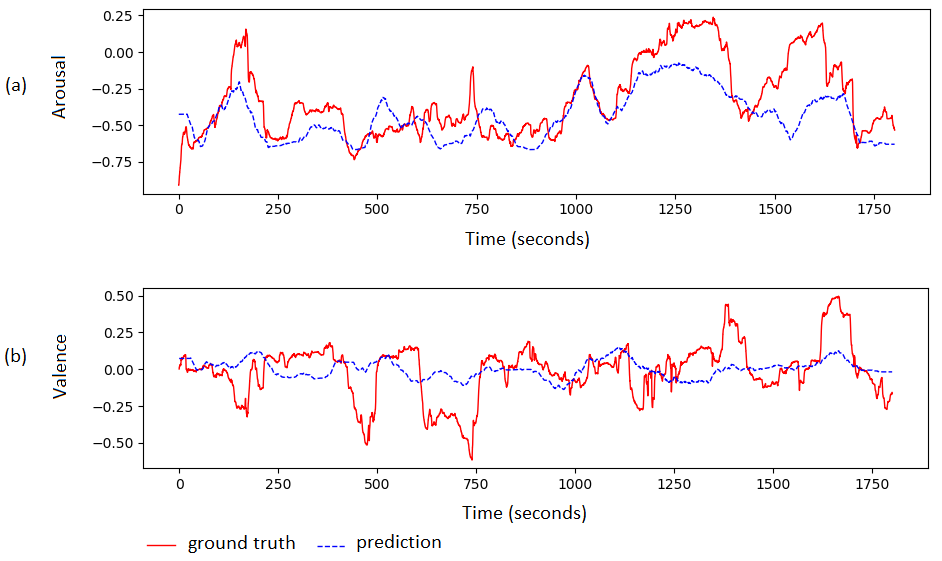}  
\end{center}
   \caption{Continuous arousal (a) and valence (b) values for experienced emotion of the ``{\em Gladiator}'' movie clip from the extended COGNIMUSE dataset.}  
\label{fig:visualization_GLA_Experienced}
\end{figure}
\begin{figure}[t!]    
\begin{center}
   \includegraphics[scale=0.5]{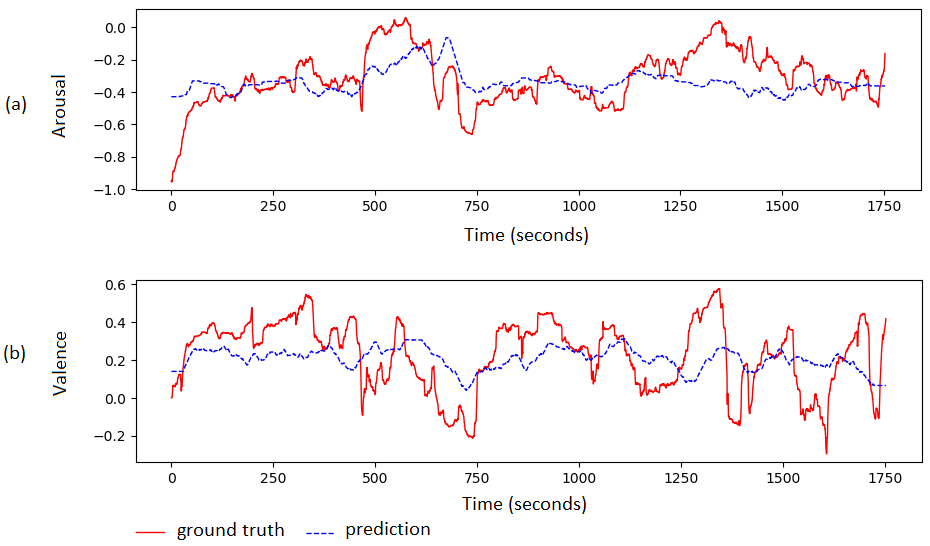}  
\end{center}
   \caption{Continuous arousal (a) and valence (b) values for experienced emotion of the ``{\em Ratatouille}'' movie clip from the extended COGNIMUSE dataset.}  
\label{fig:visualization_RAT_Experienced}
\end{figure}
%
\paragraph{Analysis of the importance of temporal memory} As described above, we propose two models, one with only fully connected layers and the other with the LSTM structure. Using a fusion of features extracted from RGB frames, optical flow, and audio, the model with fully connected layers has a higher accuracy than the LSTM approach in all predicted emotion values.  It is surprising that the LSTM does not bring any improvement in the accuracy performance. We believe this may be because in both the fully connected model and the LSTM, we use audio and optical flow features that span 400ms in the past, and this might be sufficient to carry the emotional content of the current time point.   
%
%
%
%
%
%
\begin{figure}[t!]    
\begin{center}
   \includegraphics[scale=0.5]{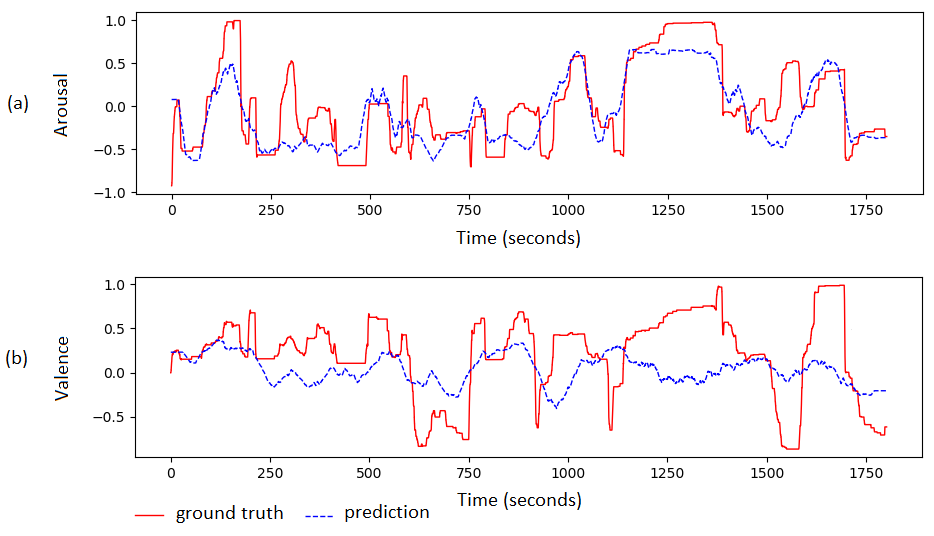}  
\end{center}
   \caption{Continuous arousal (a) and valence (b) values for
   intended emotion of the ``{\em Gladiator}'' movie clip from the extended COGNIMUSE dataset.}  
\label{fig:visualization_GLA_Intended}
\end{figure}
%
%
%
%
%
%
\begin{figure}[t!]    
\begin{center}
   \includegraphics[scale=0.5]{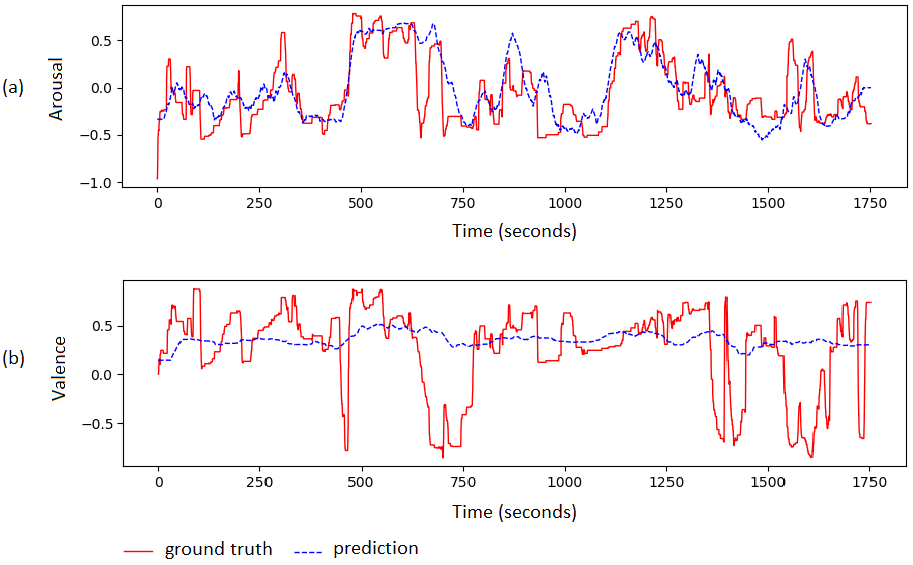}  
\end{center}
   \caption{Continuous arousal (a) and valence (b) values for intended emotion of the ``{\em Ratatouille}'' movie clip from the extended COGNIMUSE dataset.}  
\label{fig:visualization_RAT_Intended}
\end{figure}
%
%
%
\paragraph{Comparison to state-of-the-art results}
We compare our approach to~\citet{malandrakis2011supervised} for emotion prediction, in which several video, audio and music features are used as inputs of Hidden Markov Models to estimate valence and arousal values separately. Results are shown in  Tables~\ref{tab:Valence_results_Intended} and~\ref{tab:Arousal_results_Intended} for valence and arousal prediction respectively. We compare these results for intended emotion values, as previous work used these annotations to report their results. Our model outperforms the previous research, even when single feature modalities are used in our case.

We note that while we treat all videos equally,~\citet{malandrakis2011supervised} do not extract image features from cartoon movies, as they argue that the video at the image level is very different from other movies. Using optical flow and pre-trained CNNs comes with an advantage since image features capture the semantic information and the motion, regardless of the intensity, color, and if it is a cartoon or a movie with real people.
\paragraph{Visualization of the predicted Valence/Arousal values}
We visualize the ground truth and predicted continuous arousal and valence values for two movie clips namely, ``Gladiator'' (a movie with actors) and ``Ratatouille'' (an animated movie) for both experienced (Figures \ref{fig:visualization_GLA_Experienced} -  \ref{fig:visualization_RAT_Experienced})  and intended emotion (Figures~\ref{fig:visualization_GLA_Intended}-\ref{fig:visualization_RAT_Intended}) predicted by our model with fully connected layers. We observe that the arousal predictions closely match the ground truth for intended and experienced emotion for both movies. The Pearson correlation coefficients for arousal and valence are $0.77$ and $0.75$ respectively for ''Gladiator" and $0.74$ and $0.41$ respectively for ``Ratatouille''. We notice that the prediction and ground truth curves are less correlated for the valence dimension. The Pearson correlation coefficients of intended and experienced emotion in terms of arousal and valence are $0.34$ and $0.24$ respectively for ``Gladiator'', while those coefficients are $0.16$ and $0.29$ respectively for the ''Ratatouille" movie clip.  
%
%
%
%
\section{Conclusion}
In this study, we presented a multimodal approach to predict evoked/experienced emotions from videos. This approach was evaluated using both experienced and intended emotion annotations from the extended COGNIMUSE dataset. In contrast to many existing studies, we do not predict emotion from faces in the videos, but rather focus on the emotion that the film \emph{evokes} in its viewers. 

We trained multiple models, both with an without LSTM component, and evaluated their performance when using different input modalities (only audio features, only motion features, only image features) and their combinations. The resulting models show a very good performance for the audio based models, which may indicate that either the audio features are better able to capture the evoked emotion than the video features, or that audio may have a bigger influence on emotions than images. When combining all features, we are able to reach the highest performance. 

We also compared the effect of taking into consideration the sequential dependency of emotion by using an LSTM based model, with a model that does not include a temporal component but uses only fully connected layers. While both models provide high-accuracy prediction for the arousal dimension, the model with only fully connected layers achieves a significantly higher performance for the valence prediction task. In future research, this model may be further improved upon by including more audio / video features and exploring other neural network architectures. 

\subsection*{Acknowledgements}
This work was funded by the SUTD-MIT IDC grant (IDG31800103), SMART-MIT grant (ING1611118-ICT), and MOE Academic Research Fund (AcRF) Tier 2 (MOE2018-T2-2-161). H.T.P.T. was also supported by the SUTD President's Graduate Fellowship.
{\small
\bibliographystyle{IEEEtranN}
\bibliography{egbib}

\begin{thebibliography}{56}
\providecommand{\natexlab}[1]{#1}
\providecommand{\url}[1]{#1}
\csname url@samestyle\endcsname
\providecommand{\newblock}{\relax}
\providecommand{\bibinfo}[2]{#2}
\providecommand{\BIBentrySTDinterwordspacing}{\spaceskip=0pt\relax}
\providecommand{\BIBentryALTinterwordstretchfactor}{4}
\providecommand{\BIBentryALTinterwordspacing}{\spaceskip=\fontdimen2\font plus
\BIBentryALTinterwordstretchfactor\fontdimen3\font minus
  \fontdimen4\font\relax}
\providecommand{\BIBforeignlanguage}[2]{{%
\expandafter\ifx\csname l@#1\endcsname\relax
\typeout{** WARNING: IEEEtranN.bst: No hyphenation pattern has been}%
\typeout{** loaded for the language `#1'. Using the pattern for}%
\typeout{** the default language instead.}%
\else
\language=\csname l@#1\endcsname
\fi
#2}}
\providecommand{\BIBdecl}{\relax}
\BIBdecl

\bibitem[Zentner et~al.(2008)Zentner, Grandjean, and
  Scherer]{zentner2008emotions}
M.~Zentner, D.~Grandjean, and K.~R. Scherer, ``Emotions evoked by the sound of
  music: characterization, classification, and measurement.'' \emph{Emotion},
  vol.~8, no.~4, p. 494, 2008.

\bibitem[Koelsch(2010)]{koelsch2010towards}
S.~Koelsch, ``Towards a neural basis of music-evoked emotions,'' \emph{Trends
  in cognitive sciences}, vol.~14, no.~3, pp. 131--137, 2010.

\bibitem[Baumgartner et~al.(2006)Baumgartner, Esslen, and
  J{\"a}ncke]{baumgartner2006emotion}
T.~Baumgartner, M.~Esslen, and L.~J{\"a}ncke, ``From emotion perception to
  emotion experience: Emotions evoked by pictures and classical music,''
  \emph{International journal of psychophysiology}, vol.~60, no.~1, pp. 34--43,
  2006.

\bibitem[Gabrielsson(2001)]{gabrielsson2001emotion}
A.~Gabrielsson, ``Emotion perceived and emotion felt: Same or different?''
  \emph{Musicae scientiae}, vol.~5, no. 1\_suppl, pp. 123--147, 2001.

\bibitem[Kahou et~al.(2016)Kahou, Bouthillier, Lamblin, Gulcehre, Michalski,
  Konda, Jean, Froumenty, Dauphin, Boulanger-Lewandowski,
  et~al.]{kahou2016emonets}
S.~E. Kahou, X.~Bouthillier, P.~Lamblin, C.~Gulcehre, V.~Michalski, K.~Konda,
  S.~Jean, P.~Froumenty, Y.~Dauphin, N.~Boulanger-Lewandowski \emph{et~al.},
  ``Emonets: Multimodal deep learning approaches for emotion recognition in
  video,'' \emph{Journal on Multimodal User Interfaces}, vol.~10, no.~2, pp.
  99--111, 2016.

\bibitem[Hu et~al.(2017)Hu, Cai, Wang, Yao, and Chen]{hu2017learning}
P.~Hu, D.~Cai, S.~Wang, A.~Yao, and Y.~Chen, ``Learning supervised scoring
  ensemble for emotion recognition in the wild,'' in \emph{Proceedings of the
  19th ACM international conference on multimodal interaction}.\hskip 1em plus
  0.5em minus 0.4em\relax ACM, 2017, pp. 553--560.

\bibitem[Ebrahimi~Kahou et~al.(2015)Ebrahimi~Kahou, Michalski, Konda,
  Memisevic, and Pal]{ebrahimi2015recurrent}
S.~Ebrahimi~Kahou, V.~Michalski, K.~Konda, R.~Memisevic, and C.~Pal,
  ``Recurrent neural networks for emotion recognition in video,'' in
  \emph{Proceedings of the 2015 ACM on International Conference on Multimodal
  Interaction}.\hskip 1em plus 0.5em minus 0.4em\relax ACM, 2015, pp. 467--474.

\bibitem[Cowen and Keltner(2017)]{cowen2017self}
A.~S. Cowen and D.~Keltner, ``Self-report captures 27 distinct categories of
  emotion bridged by continuous gradients,'' \emph{Proceedings of the National
  Academy of Sciences}, vol. 114, no.~38, pp. E7900--E7909, 2017.

\bibitem[Picard(2000)]{picard2000affective}
R.~W. Picard, \emph{Affective computing}.\hskip 1em plus 0.5em minus
  0.4em\relax MIT press, 2000.

\bibitem[Russell(1980)]{russell1980circumplex}
J.~A. Russell, ``A circumplex model of affect.'' \emph{Journal of personality
  and social psychology}, vol.~39, no.~6, p. 1161, 1980.

\bibitem[Samara et~al.(2016)Samara, Menezes, and Galway]{samara2016feature}
A.~Samara, M.~L.~R. Menezes, and L.~Galway, ``Feature extraction for emotion
  recognition and modelling using neurophysiological data,'' in
  \emph{Ubiquitous Computing and Communications and 2016 International
  Symposium on Cyberspace and Security (IUCC-CSS), International Conference
  on}.\hskip 1em plus 0.5em minus 0.4em\relax IEEE, 2016, pp. 138--144.

\bibitem[Zlatintsi et~al.(2017)Zlatintsi, Koutras, Evangelopoulos, Malandrakis,
  Efthymiou, Pastra, Potamianos, and Maragos]{zlatintsi2017cognimuse}
A.~Zlatintsi, P.~Koutras, G.~Evangelopoulos, N.~Malandrakis, N.~Efthymiou,
  K.~Pastra, A.~Potamianos, and P.~Maragos, ``Cognimuse: a multimodal video
  database annotated with saliency, events, semantics and emotion with
  application to summarization,'' \emph{EURASIP Journal on Image and Video
  Processing}, vol. 2017, no.~1, p.~54, 2017.

\bibitem[Herremans et~al.(2017)Herremans, Yang, Chuan, Barthet, and
  Chew]{herremans2017imma}
D.~Herremans, S.~Yang, C.-H. Chuan, M.~Barthet, and E.~Chew, ``Imma-emo: A
  multimodal interface for visualising score-and audio-synchronised emotion
  annotations,'' in \emph{Proceedings of the 12th International Audio Mostly
  Conference on Augmented and Participatory Sound and Music Experiences}.\hskip
  1em plus 0.5em minus 0.4em\relax ACM, 2017, p.~11.

\bibitem[Lang(1995)]{lang1995emotion}
P.~J. Lang, ``The emotion probe: studies of motivation and attention.''
  \emph{American psychologist}, vol.~50, no.~5, p. 372, 1995.

\bibitem[Krizhevsky et~al.(2012)Krizhevsky, Sutskever, and
  Hinton]{krizhevsky2012imagenet}
A.~Krizhevsky, I.~Sutskever, and G.~E. Hinton, ``Imagenet classification with
  deep convolutional neural networks,'' in \emph{Advances in neural information
  processing systems}, 2012, pp. 1097--1105.

\bibitem[Simonyan and Zisserman(2014{\natexlab{a}})]{simonyan2014very}
K.~Simonyan and A.~Zisserman, ``Very deep convolutional networks for
  large-scale image recognition,'' \emph{arXiv preprint arXiv:1409.1556}, 2014.

\bibitem[He et~al.(2016)He, Zhang, Ren, and Sun]{he2016deep}
K.~He, X.~Zhang, S.~Ren, and J.~Sun, ``Deep residual learning for image
  recognition,'' in \emph{Proceedings of the IEEE conference on computer vision
  and pattern recognition}, 2016, pp. 770--778.

\bibitem[Szegedy et~al.(2015)Szegedy, Liu, Jia, Sermanet, Reed, Anguelov,
  Erhan, Vanhoucke, and Rabinovich]{szegedy2015going}
C.~Szegedy, W.~Liu, Y.~Jia, P.~Sermanet, S.~Reed, D.~Anguelov, D.~Erhan,
  V.~Vanhoucke, and A.~Rabinovich, ``Going deeper with convolutions,'' in
  \emph{Proceedings of the IEEE conference on computer vision and pattern
  recognition}, 2015, pp. 1--9.

\bibitem[Ji et~al.(2012)Ji, Xu, Yang, and Yu]{ji20123d}
S.~Ji, W.~Xu, M.~Yang, and K.~Yu, ``3d convolutional neural networks for human
  action recognition,'' \emph{IEEE transactions on pattern analysis and machine
  intelligence}, vol.~35, no.~1, pp. 221--231, 2012.

\bibitem[Simonyan and Zisserman(2014{\natexlab{b}})]{simonyan2014two}
K.~Simonyan and A.~Zisserman, ``Two-stream convolutional networks for action
  recognition in videos,'' in \emph{Advances in neural information processing
  systems}, 2014, pp. 568--576.

\bibitem[Carreira and Zisserman(2017)]{carreira2017quo}
J.~Carreira and A.~Zisserman, ``Quo vadis, action recognition? a new model and
  the kinetics dataset,'' in \emph{proceedings of the IEEE Conference on
  Computer Vision and Pattern Recognition}, 2017, pp. 6299--6308.

\bibitem[Eyben(2016)]{EybenOpenSMILE}
F.~Eyben, \emph{Real-time Speech and Music Classification by Large Audio
  Feature Space Extraction}, 1st~ed.\hskip 1em plus 0.5em minus 0.4em\relax
  Springer Publishing Company, Incorporated, 2016.

\bibitem[Fan et~al.(2016)Fan, Lu, Li, and Liu]{fan2016video}
Y.~Fan, X.~Lu, D.~Li, and Y.~Liu, ``Video-based emotion recognition using
  cnn-rnn and c3d hybrid networks,'' in \emph{Proceedings of the 18th ACM
  International Conference on Multimodal Interaction}.\hskip 1em plus 0.5em
  minus 0.4em\relax ACM, 2016, pp. 445--450.

\bibitem[Pini et~al.(2017)Pini, Ahmed, Cornia, Baraldi, Cucchiara, and
  Huet]{pini2017modeling}
S.~Pini, O.~B. Ahmed, M.~Cornia, L.~Baraldi, R.~Cucchiara, and B.~Huet,
  ``Modeling multimodal cues in a deep learning-based framework for emotion
  recognition in the wild,'' in \emph{Proceedings of the 19th ACM International
  Conference on Multimodal Interaction}.\hskip 1em plus 0.5em minus 0.4em\relax
  ACM, 2017, pp. 536--543.

\bibitem[Malandrakis et~al.(2011)Malandrakis, Potamianos, Evangelopoulos, and
  Zlatintsi]{malandrakis2011supervised}
N.~Malandrakis, A.~Potamianos, G.~Evangelopoulos, and A.~Zlatintsi, ``A
  supervised approach to movie emotion tracking,'' in \emph{Acoustics, Speech
  and Signal Processing (ICASSP), 2011 IEEE International Conference on}.\hskip
  1em plus 0.5em minus 0.4em\relax IEEE, 2011, pp. 2376--2379.

\bibitem[Evangelopoulos et~al.(2013)Evangelopoulos, Zlatintsi, Potamianos,
  Maragos, Rapantzikos, Skoumas, and Avrithis]{evangelopoulos2013multimodal}
G.~Evangelopoulos, A.~Zlatintsi, A.~Potamianos, P.~Maragos, K.~Rapantzikos,
  G.~Skoumas, and Y.~Avrithis, ``Multimodal saliency and fusion for movie
  summarization based on aural, visual, and textual attention,'' \emph{IEEE
  Transactions on Multimedia}, vol.~15, no.~7, pp. 1553--1568, 2013.

\bibitem[Kanade(2005)]{Kanade:2005:FEA:2101315.2101317}
\BIBentryALTinterwordspacing
T.~Kanade, ``Facial expression analysis,'' in \emph{Proceedings of the Second
  International Conference on Analysis and Modelling of Faces and Gestures},
  ser. AMFG'05.\hskip 1em plus 0.5em minus 0.4em\relax Berlin, Heidelberg:
  Springer-Verlag, 2005, pp. 1--1. [Online]. Available:
  \url{http://dx.doi.org/10.1007/11564386_1}
\BIBentrySTDinterwordspacing

\bibitem[Cohn and Torre(2014)]{Cohn_automatedface}
J.~F. Cohn and F.~D.~L. Torre, ``Automated face analysis for affective
  computing,'' 2014.

\bibitem[Levi and Hassner(2015)]{levi2015emotion}
G.~Levi and T.~Hassner, ``Emotion recognition in the wild via convolutional
  neural networks and mapped binary patterns,'' in \emph{Proceedings of the
  2015 ACM on international conference on multimodal interaction}.\hskip 1em
  plus 0.5em minus 0.4em\relax ACM, 2015, pp. 503--510.

\bibitem[Kaya et~al.(2017)Kaya, G{\"u}rp{\i}nar, and Salah]{kaya2017video}
H.~Kaya, F.~G{\"u}rp{\i}nar, and A.~A. Salah, ``Video-based emotion recognition
  in the wild using deep transfer learning and score fusion,'' \emph{Image and
  Vision Computing}, vol.~65, pp. 66--75, 2017.

\bibitem[Zheng et~al.(2018)Zheng, Cao, Chen, and Xu]{zheng2018multimodal}
Z.~Zheng, C.~Cao, X.~Chen, and G.~Xu, ``Multimodal emotion recognition for
  one-minute-gradual emotion challenge,'' \emph{arXiv preprint
  arXiv:1805.01060}, 2018.

\bibitem[Szegedy et~al.(2017)Szegedy, Ioffe, Vanhoucke, and
  Alemi]{szegedy2017inception}
C.~Szegedy, S.~Ioffe, V.~Vanhoucke, and A.~A. Alemi, ``Inception-v4,
  inception-resnet and the impact of residual connections on learning,'' in
  \emph{Thirty-First AAAI Conference on Artificial Intelligence}, 2017.

\bibitem[Mase(1991)]{mase1991recognition}
K.~Mase, ``Recognition of facial expression from optical flow,'' \emph{IEICE
  TRANSACTIONS on Information and Systems}, vol.~74, no.~10, pp. 3474--3483,
  1991.

\bibitem[Wang et~al.(2015)Wang, Xiong, Wang, and Qiao]{wang2015towards}
L.~Wang, Y.~Xiong, Z.~Wang, and Y.~Qiao, ``Towards good practices for very deep
  two-stream convnets,'' \emph{arXiv preprint arXiv:1507.02159}, 2015.

\bibitem[B.~Meyer(2008)]{Meyer56}
L.~B.~Meyer, ``Emotion and meaning in music,'' \emph{Journal of Music Theory},
  vol.~16, 01 2008.

\bibitem[Panksepp and Bernatzky(2002)]{panksepp2002emotional}
J.~Panksepp and G.~Bernatzky, ``Emotional sounds and the brain: the
  neuro-affective foundations of musical appreciation,'' \emph{Behavioural
  processes}, vol.~60, no.~2, pp. 133--155, 2002.

\bibitem[Doughty et~al.(2016)Doughty, Duffy, and Harada]{doughty2016practices}
K.~Doughty, M.~Duffy, and T.~Harada, ``Practices of emotional and affective
  geographies of sound,'' 2016.

\bibitem[Herremans and Chew(2016)]{herremans2016}
D.~Herremans and E.~Chew, ``{Tension ribbons: Quantifying and visualising tonal
  tension},'' in \emph{{Second International Conference on Technologies for
  Music Notation and Representation (TENOR)}}, Cambridge, UK, 05/2016 2016.

\bibitem[Mathieu et~al.(2010)Mathieu, Essid, Fillon, Prado, and
  Richard]{mathieu2010yaafe}
B.~Mathieu, S.~Essid, T.~Fillon, J.~Prado, and G.~Richard, ``Yaafe, an easy to
  use and efficient audio feature extraction software.'' in \emph{ISMIR}, 2010,
  pp. 441--446.

\bibitem[Aytar et~al.(2016)Aytar, Vondrick, and Torralba]{aytar2016soundnet}
Y.~Aytar, C.~Vondrick, and A.~Torralba, ``Soundnet: Learning sound
  representations from unlabeled video,'' in \emph{Advances in neural
  information processing systems}, 2016, pp. 892--900.

\bibitem[Hershey et~al.(2017)Hershey, Chaudhuri, Ellis, Gemmeke, Jansen, Moore,
  Plakal, Platt, Saurous, Seybold, et~al.]{hershey2017cnn}
S.~Hershey, S.~Chaudhuri, D.~P. Ellis, J.~F. Gemmeke, A.~Jansen, R.~C. Moore,
  M.~Plakal, D.~Platt, R.~A. Saurous, B.~Seybold \emph{et~al.}, ``Cnn
  architectures for large-scale audio classification,'' in \emph{2017 ieee
  international conference on acoustics, speech and signal processing
  (icassp)}.\hskip 1em plus 0.5em minus 0.4em\relax IEEE, 2017, pp. 131--135.

\bibitem[Hochreiter and Schmidhuber(1997)]{hochreiter1997long}
S.~Hochreiter and J.~Schmidhuber, ``Long short-term memory,'' \emph{Neural
  computation}, vol.~9, no.~8, pp. 1735--1780, 1997.

\bibitem[Graves(2013)]{graves2013generating}
A.~Graves, ``Generating sequences with recurrent neural networks,'' \emph{arXiv
  preprint arXiv:1308.0850}, 2013.

\bibitem[Zaremba and Sutskever(2014)]{zaremba2014learning}
W.~Zaremba and I.~Sutskever, ``Learning to execute,'' \emph{arXiv preprint
  arXiv:1410.4615}, 2014.

\bibitem[Goyal et~al.(2016)Goyal, Kumar, Guha, and
  Narayanan]{goyal2016multimodal}
A.~Goyal, N.~Kumar, T.~Guha, and S.~S. Narayanan, ``A multimodal
  mixture-of-experts model for dynamic emotion prediction in movies,'' in
  \emph{2016 IEEE International Conference on Acoustics, Speech and Signal
  Processing (ICASSP)}.\hskip 1em plus 0.5em minus 0.4em\relax IEEE, 2016, pp.
  2822--2826.

\bibitem[Sivaprasad et~al.(2018)Sivaprasad, Joshi, Agrawal, and
  Pedanekar]{sivaprasad2018multimodal}
S.~Sivaprasad, T.~Joshi, R.~Agrawal, and N.~Pedanekar, ``Multimodal continuous
  prediction of emotions in movies using long short-term memory networks,'' in
  \emph{Proceedings of the 2018 ACM on International Conference on Multimedia
  Retrieval}.\hskip 1em plus 0.5em minus 0.4em\relax ACM, 2018, pp. 413--419.

\bibitem[Sun et~al.(2018)Sun, Yang, Liu, and Kautz]{sun2018pwc}
D.~Sun, X.~Yang, M.-Y. Liu, and J.~Kautz, ``Pwc-net: Cnns for optical flow
  using pyramid, warping, and cost volume,'' in \emph{Proceedings of the IEEE
  Conference on Computer Vision and Pattern Recognition}, 2018, pp. 8934--8943.

\bibitem[Butler et~al.(2012)Butler, Wulff, Stanley, and
  Black]{butler2012naturalistic}
D.~J. Butler, J.~Wulff, G.~B. Stanley, and M.~J. Black, ``A naturalistic open
  source movie for optical flow evaluation,'' in \emph{European conference on
  computer vision}.\hskip 1em plus 0.5em minus 0.4em\relax Springer, 2012, pp.
  611--625.

\bibitem[Ilg et~al.(2017)Ilg, Mayer, Saikia, Keuper, Dosovitskiy, and
  Brox]{ilg2017flownet}
E.~Ilg, N.~Mayer, T.~Saikia, M.~Keuper, A.~Dosovitskiy, and T.~Brox, ``Flownet
  2.0: Evolution of optical flow estimation with deep networks,'' in \emph{IEEE
  conference on computer vision and pattern recognition (CVPR)}, vol.~2, 2017,
  p.~6.

\bibitem[Ranjan and Black(2017)]{ranjan2017optical}
A.~Ranjan and M.~J. Black, ``Optical flow estimation using a spatial pyramid
  network,'' in \emph{Proceedings of the IEEE Conference on Computer Vision and
  Pattern Recognition}, 2017, pp. 4161--4170.

\bibitem[Xu et~al.(2017)Xu, Ranftl, and Koltun]{xu2017accurate}
J.~Xu, R.~Ranftl, and V.~Koltun, ``Accurate optical flow via direct cost volume
  processing,'' in \emph{Proceedings of the IEEE Conference on Computer Vision
  and Pattern Recognition}, 2017, pp. 1289--1297.

\bibitem[Bailer et~al.(2015)Bailer, Taetz, and Stricker]{bailer2015flow}
C.~Bailer, B.~Taetz, and D.~Stricker, ``Flow fields: Dense correspondence
  fields for highly accurate large displacement optical flow estimation,'' in
  \emph{Proceedings of the IEEE international conference on computer vision},
  2015, pp. 4015--4023.

\bibitem[Schuller et~al.(2010)Schuller, Steidl, Batliner, Burkhardt, Devillers,
  M{\"u}ller, and Narayanan]{schuller2010interspeech}
B.~Schuller, S.~Steidl, A.~Batliner, F.~Burkhardt, L.~Devillers, C.~M{\"u}ller,
  and S.~S. Narayanan, ``The interspeech 2010 paralinguistic challenge,'' in
  \emph{Eleventh Annual Conference of the International Speech Communication
  Association}, 2010.

\bibitem[Eyben et~al.(2016)Eyben, Weninger, W{\"o}llmer, and
  Shuller]{eyben2016open}
F.~Eyben, F.~Weninger, M.~W{\"o}llmer, and B.~Shuller, ``Open-source media
  interpretation by large feature-space extraction,'' 2016.

\bibitem[Donahue et~al.(2015)Donahue, Anne~Hendricks, Guadarrama, Rohrbach,
  Venugopalan, Saenko, and Darrell]{donahue2015long}
J.~Donahue, L.~Anne~Hendricks, S.~Guadarrama, M.~Rohrbach, S.~Venugopalan,
  K.~Saenko, and T.~Darrell, ``Long-term recurrent convolutional networks for
  visual recognition and description,'' in \emph{Proceedings of the IEEE
  conference on computer vision and pattern recognition}, 2015, pp. 2625--2634.

\bibitem[Savitzky and Golay(1964)]{savitzky1964smoothing}
A.~Savitzky and M.~J. Golay, ``Smoothing and differentiation of data by
  simplified least squares procedures.'' \emph{Analytical chemistry}, vol.~36,
  no.~8, pp. 1627--1639, 1964.

\end{thebibliography}
}
\end{document}